\pdfoutput=1

\documentclass[11pt]{article}
\usepackage{amssymb}
\usepackage{graphicx}
\usepackage{booktabs}
\usepackage{tabularx}
\usepackage{array}
\usepackage{xcolor}
\usepackage{colortbl}
\usepackage{paralist}
\usepackage{listings}
\usepackage{enumitem}
\usepackage{makecell}
\usepackage{amsmath}
\usepackage[final]{acl}

\usepackage{times}
\usepackage{latexsym}
\usepackage{tcolorbox}

\usepackage[T1]{fontenc}

\usepackage[utf8]{inputenc}

\usepackage{microtype}

\usepackage{inconsolata}

\newcommand{\methodAbbrev}{ARTER~}

\title{Leveraging the Power of Large Language Models in Entity Linking via Adaptive Routing and Targeted Reasoning 
}

\author{
Yajie Li\textsuperscript{1}, Albert Galimov\textsuperscript{1}, Mitra Datta Ganapaneni\textsuperscript{1}, Pujitha Thejaswi\textsuperscript{1} \\
\textbf{De Meng\textsuperscript{2}, Priyanshu Kumar\textsuperscript{2}, Saloni Potdar\textsuperscript{2}} \\
\textsuperscript{1}College of Information and Computer Sciences, University of Massachusetts Amherst \\
\textsuperscript{2} Apple \\
\texttt{\{yajieli, agalimov, mganapaneni, pthejaswi\}@umass.edu}, \\
\texttt{\{de\_meng, priyanshu\_kumar, s\_potdar\}@apple.com} \\
}

\begin{document}
\maketitle
\begin{abstract}
Entity Linking (EL) has traditionally relied on large annotated datasets and extensive model fine-tuning. While recent few-shot methods leverage large language models (LLMs) through prompting to reduce training requirements, they often suffer from inefficiencies due to expensive LLM-based reasoning. \methodAbbrev (Adaptive Routing and Targeted Entity Reasoning) presents a structured pipeline that achieves high performance without deep fine-tuning by strategically combining candidate generation, context-based scoring, adaptive routing, and selective reasoning. \methodAbbrev computes a small set of complementary signals (both embedding and LLM-based) over the retrieved candidates to categorize contextual mentions into \textit{easy} and \textit{hard} cases. The cases are then handled by a low-computational entity linker (e.g. ReFinED) and more expensive targeted LLM-based reasoning respectively. On standard benchmarks, \methodAbbrev outperforms ReFinED by up to +4.47\%, with an average gain of +2.53\% on 5 out of 6 datasets, and performs comparably to pipelines using LLM-based reasoning for all mentions, while being as twice as efficient in terms of the number of LLM tokens.

\end{abstract}

\section{Introduction}

Entity Linking (EL), also known as Named Entity Disambiguation (NED), is the critical process of accurately associating ambiguous textual mentions with their corresponding specific entities within a knowledge base. 
Entity linking systems are widely used in search engines, automated knowledge extraction platforms, question-answering systems, and other large-scale NLP systems.


Traditional EL methods depend on large labeled datasets and extensive fine-tuning, which limits their adaptability to new domains and their ability to scale \cite{ayoola-etal-2022-refined, wu2019zero}. They also struggle with “hard cases” including: 1) low-context mentions, which have insufficient context to disambiguate (e.g., “Ireland” in a short snippet), 2) lexical ambiguity, when the same surface form maps to multiple entity types or senses (e.g., a country versus its national team), and 3) knowledge-intensive cases, which demand external or implicit world knowledge for correct linking (e.g., interpreting ‘the Big Apple’ as New York City involves background knowledge rather than textual clues).

Large Language Models (LLMs) have transformed entity linking by using prompt-based methods \citep{xin2024llmaellargelanguagemodels} in place of large annotated datasets thus improving performance on difficult disambiguation cases through their advanced reasoning. However, current LLM-centric approaches are inefficient, as they require repeated model calls at each step, leading to increased cost and slower processing.

Traditional EL systems struggle to handle cases where mention complexity varies, as they do not adapt based on varying complexities of mentions. Our proposed framework, \methodAbbrev (Adaptive Routing and Targeted Entity Reasoning), dynamically applies advanced reasoning based on mention complexity—offloading only the most ambiguous cases to an LLM, which resolves them using a targeted reasoning prompt, thus eliminating the need of finetuning or annotating new examples for those hard mentions. 

\begin{figure*}[h!]
    \centering
    \includegraphics[width=0.95\textwidth]{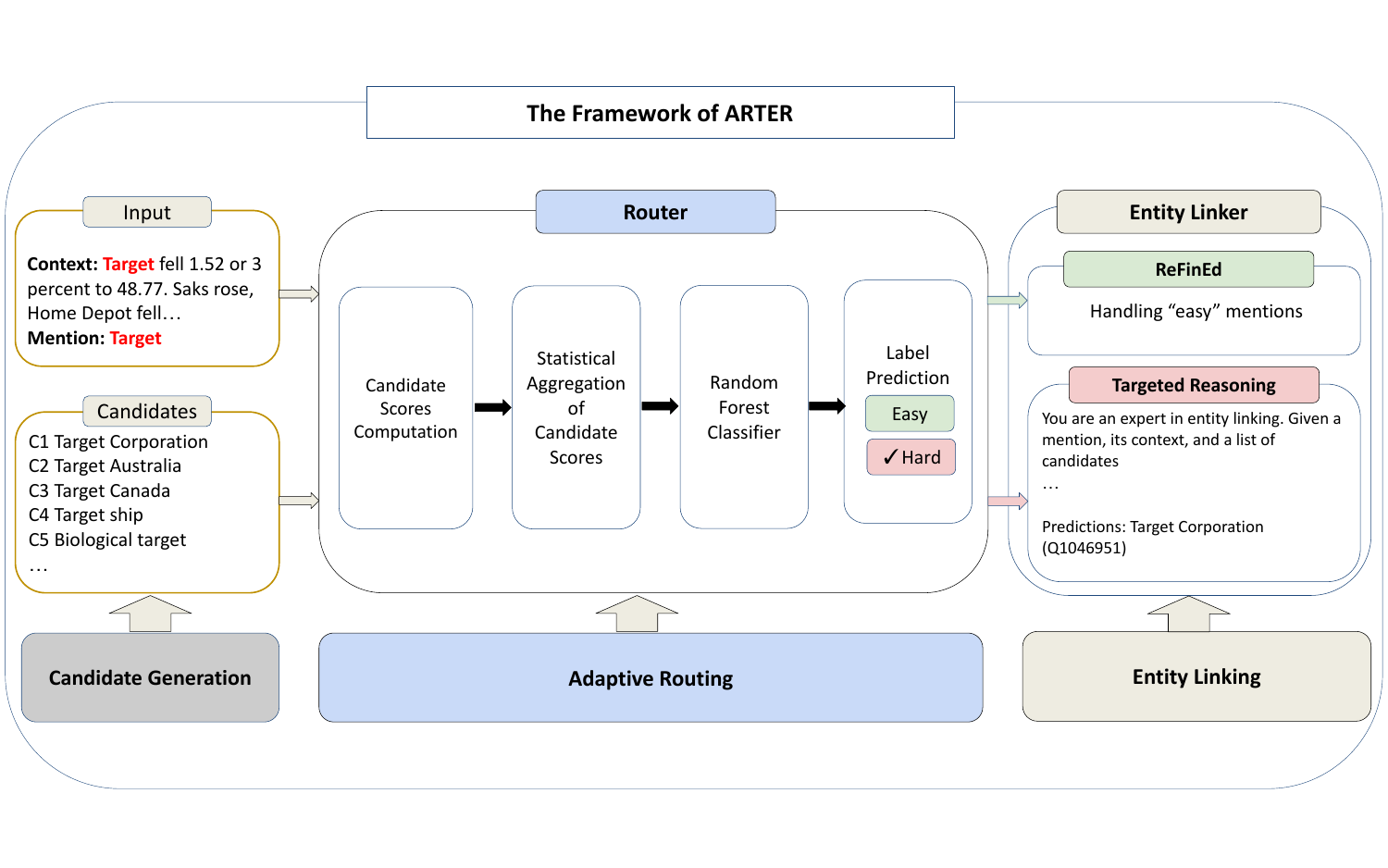}
    \caption{\methodAbbrev leverages a light-weight router on top of the given context and retrieved candidate entities to categorize contextual mentions into \textit{easy} and \textit{hard} cases. The cases are then handled by a low-computational entity linker (e.g. ReFinED) and more expensive targeted LLM-based reasoning respectively.}
    \label{fig:pipeline}
\end{figure*}

\methodAbbrev is a structured pipeline as illustrated in Figure ~\ref{fig:pipeline}. The pipeline starts with generating candidate entities and computing candidate scores; then using these scoring signals, a lightweight router classifies mentions as easy or hard cases. Easy mentions are handled by fast traditional EL models, while hard cases are routed through a reasoning prompt for refined disambiguation. This design offers two key benefits. First, it avoids the computational cost of full-model fine-tuning required by systems like ReFinED, which typically need to be adapted to new datasets or domains to achieve strong performance. In our approach, ReFinED remains frozen, serving for fast high-recall candidate generation and handling “easy” mentions. Second, by applying targeted reasoning prompts to difficult mentions, our method avoids relying solely on prompt-based inference as in OneNet \cite{liu-etal-2024-onenet} and is more efficient as only hard cases incur the LLM cost rather than every mention.
We evaluate \methodAbbrev on standard benchmarks—AIDA \cite{hoffart-etal-2011-robust}, MSNBC \cite{cucerzan-2007-large}, ACE2004 \cite{ratinov-etal-2011-local}, AQUAINT \cite{milne2008learning}, CWEB \citep{gabrilovich2013facc1}, and WIKI \cite{guo2018robust_preprint} and demonstrate improved performance through efficient, selective use of LLM reasoning. Based on our benchmarking results, we try to answer the following research questions:

\begin{enumerate}
  \item Can the strategic application of LLM-based techniques within \methodAbbrev significantly improve EL performance for ambiguous mentions (i.e., “hard cases” where traditional models typically falter)?
  
  \item Do all mentions require LLM reasoning? If not, which mentions can be handled by traditional models? How can we reduce the cost of LLM-based reasoning without sacrificing accuracy?

  \item How does \methodAbbrev compare to the full prompting approach in terms of accuracy and computational efficiency?

\end{enumerate}

\section{Related Work}


Entity Linking (EL) has seen significant progress with systems like BLINK and ReFinED. \citet{wu-etal-2020-scalable} proposed BLINK, a zero-shot framework that uses a bi-encoder for candidate retrieval and a cross-encoder for ranking. While effective on large datasets such as TACKBP-2010, BLINK struggles with long-tail entities due to limited contextual understanding. ReFinED \citep{ayoola-etal-2022-refined} addresses efficiency by processing all mentions in a single pass and improves accuracy with fine-grained entity typing and entity priors. Our work builds on these systems by using its candidate generation module to obtain high-quality entity candidates, while avoiding the burden of fine-tuning by relying on prompt-based strategies instead. 

Despite our focus on EL, recent progress in Relation Extraction (RE) is closely related because both tasks require understanding and reasoning over entity–context relationships. Many prompting techniques developed for RE, such as few-shot and chain-of-thought reasoning, have proven transferable to EL and help LLMs perform contextual disambiguation. Recent work explores LLM-based approaches to both EL and RE. \citet{wadhwa2021revisiting} show that few-shot prompting can match RE baselines, while Chain-of-Thought (CoT) improves accuracy. \citet{entity-disambiguation-fusion-decoding} apply retrieval-augmentation for entity disambiguation, but find performance still lags behind SOTA on GERBIL \citep{hoffart2015gerbil}. LLMAEL \citep{xin2024llmaellargelanguagemodels} augments mention context with LLM-generated descriptions but requires fine-tuning. OneNet \citep{liu-etal-2024-onenet} uses a fully prompt-based EL pipeline without retraining, yet lacks adaptive control over reasoning depth.

\section{Method}

\subsection{Problem Formulation}
Given the context (e.g. an article or piece of text) of a mention, the task of entity linking is to correctly identify the candidate entity associated with the mention as per a knowledge base. 

\subsection{Candidate Generation}


Our entity-linking pipeline leverages the ReFinED candidate-generation module, which returns an initial pool of 30 ranked candidates for each mention.
Each candidate refers to a unique entity entry (article) in the underlying English Wikipedia knowledge base, which serves as the reference set for disambiguation. For additional metadata such as entity titles and short descriptions, we retrieve the corresponding information from Wikidata linked to each Wikipedia entity. These enriched representations are then used in the subsequent routing and reasoning stages.

\subsection{Router}
For efficiency, we retain the top 10 from the 30 retrieved candidates to train the router. For these top 10 candidates, we compute three cosine-similarity scores using a Sentence Transformer encoder over each candidate’s title+description. Let $\mathbf{c}\in\mathbb{R}^d$ denote the vector for the mention’s surrounding context, $\mathbf{m}\in\mathbb{R}^d$ the vector for the mention surface form, and $\mathbf{e}_i\in\mathbb{R}^d$ the vector for the $i$-th candidate entity (title+description). Cosine similarity between any two vectors $\mathbf{u},\mathbf{v}\in\mathbb{R}^d$ is
\begin{equation}
\operatorname{cos}(\mathbf{u},\mathbf{v}) \;=\; \frac{\mathbf{u} \cdot \mathbf{v}}{\lVert \mathbf{u}\rVert \,\lVert \mathbf{v}\rVert }.
\end{equation}
For each candidate $i$, we define:
\begin{equation}
\theta_{1}(i)=\operatorname{cos}\!\big(\mathbf{c},\,\mathbf{e}_i\big),
\end{equation}
\begin{equation}
\theta_{2}(i)=\operatorname{cos}\!\big(\mathbf{m},\,\mathbf{e}_i\big),
\end{equation}
\begin{equation}
\theta_{3}(i)=\max_{i\neq j}\operatorname{cos}\!\big(\mathbf{e}_i,\,\mathbf{e}_j\big).
\end{equation}
where 
$\theta_{1}(i)$ is the context–entity similarity, $\theta_{2}(i)$ is the mention–entity similarity, and $\theta_{3}(i)$ is the inter-candidate similarity (the maximum similarity between candidate $i$ and any other candidate $i\neq j$).

These similarity measures are computed in parallel with a fourth score generated by a small, non-reasoning language model (e.g., Llama-3.1-8B-Instruct). The LLM receives a single-turn prompt containing the mention, its context, and the top-10 candidates (with titles + descriptions) and outputs a confidence score $\phi(i)$ for each candidate. All four scores are passed to our router, which uses them to decide whether heavy reasoning via the LLM is necessary or if the mention can be linked using lower-cost methods (see Appendix~\ref{sec:appendix} for the full LLM prompt template).

Since any single sample can produce multiple candidate entities—each with four confidence scores—we aggregate all candidate scores into a set of statistical features as shown in Table \ref{tab:feature_summary}.

\vspace{-\parskip}
To distinguish between “easy” and “hard” cases, we use the predictions from the ReFinED end-to-end entity-linking model. Specifically, for each mention in the AIDA training set (18,395 instances), we obtain the predicted entity from ReFinED. If the predicted entity matches the gold-standard label, we mark the case as \textit{easy}; if the prediction is incorrect (i.e., the predicted entity does not align with the gold label), we mark the case as \textit{hard}.

The predictions from ReFinED on the AIDA training set are used to generate “easy” and “hard” labels, which serve as training data for a Random Forest classifier. This classifier is then applied to test instances, splitting them into two subsets: those predicted as easy are handled directly by the ReFinED module, while those predicted as hard are routed to the reasoning module with an advanced prompt.



To further illustrate the separation between easy and hard cases, we provide representative examples below:

\begin{tcolorbox}[title=Easy Linking Sample,fontupper=\tiny, fontlower=\tiny]
\begin{verbatim}
Easy Case: Washington, D.C.
Mention: Washington
Context: Documented From NPR news in Washington I’m Corey Flintoff...
Correct Entity: Washington, D.C. (Q61) 
----------------------------------------------------------------
Explanation: The mention appears in a well-known journalistic context
(NPR), and ReFinED confidently disambiguates it using surface-level 
cues, without needing additional reasoning.
\end{verbatim}
\end{tcolorbox}



\begin{tcolorbox}[title=Hard Linking Sample,fontupper=\tiny, fontlower=\tiny]
\begin{verbatim}
Mention: Target
Context: Target fell 1.52 or 3 percent to 48.77. Saks rose, 
Home Depot fell...
Correct Entity: Target Corporation (Q1046951)
----------------------------------------------------------------
Explanation: The context involves financial information and lists 
multiple retail companies, suggesting a stock market setting. 
However, "Target" is a highly ambiguous mention with many possible 
entity candidates, including Target Corporation, Target Australia, 
Target Canada, and other non-retail meanings. 
\end{verbatim}
\end{tcolorbox}


\begin{table}[t]
  \centering
  \renewcommand{\arraystretch}{0.9}
  \setlength{\tabcolsep}{6pt}
  \footnotesize
  \begin{tabular}{@{} l  p{0.70\columnwidth} @{}}
    \toprule
    \textbf{Feature}   & \textbf{Description} \\
    \midrule
    top1               & Highest candidate score after similarity-penalization\textsuperscript{1} \\
    top2               & Second-highest candidate score after similarity-penalization\textsuperscript{1} \\
    margin             & Difference between \texttt{top1} and \texttt{top2} (i.e.\ $\text{top1}-\text{top2}$) \\
    entropy            & Shannon entropy (base-2) of the flattened candidate-score distribution \\
    n\_cands           & Number of entity candidates retrieved for the mention \\
    sent\_len          & Length (in tokens) of the sentence containing the mention \\
    score\_1           & Average of score $\theta_{1}$ across all candidates \\
    score\_2           & Average of score $\theta_{2}$ across all candidates \\
    score\_3           & Average of score $\theta_{3}$ across all candidates \\
    score\_4           & Average of score $\phi$ across all candidates \\
    \bottomrule
  \end{tabular}
  \caption{Summary of the ten features used to decide whether a mention is “easy” or “hard.”  
    \textsuperscript{1}Both \texttt{top1} and \texttt{top2} are computed as 
    \(\displaystyle \frac{\theta_{1} + \theta_{2} - \theta_{3} + \phi}{3}\), 
    where $\theta_{3}$ is subtracted to penalize similarity among candidates.}
  \label{tab:feature_summary}
\end{table}


\subsection{Reasoning Module}

For hard cases, a Reasoning Module is leveraged to integrate entity description generation and targeted reasoning. For each mention, the LLM-based reasoning is formulated as a multiple-choice task over the top 30 candidates retrieved by ReFinED’s candidate generation module. We study various reasoning models, including Deepseek R1 0528, GPT-4.1, Claude 3 Haiku and Claude 3.5 Sonnet. We include the results of the best reasoning models in Table \ref{tab:hybrid-comparison}. 

\subsubsection{Prompting Strategies}

To fully utilize the reasoning capabilities of LLMs, we experimented with various prompting strategies, including zero-shot \citep{kojima2023zeroshot}, few-shot \citep{brown2020fewshot}, Chain-of-Thought (CoT) \citep{wei2023cot}, contrastive prompting \citep{chia2023contrastive}, and ReAct-style prompting \citep{yao2023reactsynergizingreasoningacting}. Contrastive prompting enhances robustness by providing positive and negative examples, while ReAct-style prompting interleaves reasoning with actions, enabling step-by-step thinking and tool or information access. A detailed comparison of all strategies on the ACE2004 hard cases dataset is shown in Table~\ref{tab:prompt-strategy-accuracy}. The compasison shows that the most effective approach turns out to be the one combining few-shot examples with CoT reasoning. This method encourages the model to analyze mentions and candidates step by step, leveraging entity descriptions and category consistency. The prompt format explicitly aligns each mention and its surrounding context with a list of candidate entities and requires the model to identify the most semantically compatible option or return \(-1\) if none fits. Few-shot examples illustrate the step-by-step reasoning process that links contextual cues to the appropriate entity description, grounding predictions in the candidate information provided.


The full prompt template, including system instructions, few-shot examples, and the output format, is provided in Table \ref{tab:module_prompts_combined} (Appendix~\ref{sec:appendix}). A sample output generated from the reasoning module is shown below to demonstrate the reasoning output (the candidate entities are not fully shown due to space limit):

\begin{tcolorbox}[title=Reasoning Sample Generation,fontupper=\tiny, fontlower=\tiny]
\begin{verbatim}
Mention: Target
Context: Target fell 1.52 or 3 percent to 48.77. Saks rose, 
Home Depot fell...
Candidate Entities (subset):
  - Target Corporation (Q1046951): American retailer 
    and supermarket chain
  - Target Australia (Q7685854): Australian department 
    store chain
  - Target Canada (Q4140889): Canadian subsidiary 
    of Target Corporation
  - Target (Q180238): British television series
  - Target Books (Q12005242): Publisher of Doctor 
    Who novelizations
  - Target (Q971985): 1985 film by Arthur Penn
  - Biological target (Q904407): Molecular target in 
    drug design
  - ... 
----------------------------------------------------------------
Prediction: Target Corporation (Q1046951)
LLM Reasoning:
1. The context is discussing stock market performance and mentions 
various companies and their stock prices.  
2. 'Target' is mentioned alongside other retail companies 
like Home Depot and Saks.  
3. The text states 'Target fell 1.52 or 3 percent to 48.77', 
which refers to stock price movement.  
4. Among the candidates, only Target Corporation 
(an American retailer) fits this context as a publicly 
traded company.  
5. Other candidates like Target Australia or Target 
Canada are not likely to be traded on the US stock market.  
6. The remaining candidates are unrelated to retail or 
the stock market. Therefore, Target Corporation is the 
most appropriate match for this mention.
\end{verbatim}
\end{tcolorbox}




\section{Experimental setup}

\subsection{Datasets}

\paragraph{Data source}
A comprehensive and diverse evaluation was conducted on publicly available datasets. These datasets cover a wide range of domains by including both standard benchmarks and challenging cases, like long-tail entity linking. The datasets used to benchmark performance are the following: AIDA \cite{hoffart-etal-2011-robust}, MSNBC \cite{cucerzan-2007-large}, ACE2004 \cite{ratinov-etal-2011-local}, AQUAINT \cite{milne2008learning}, CWEB \citep{gabrilovich2013facc1} and WIKI \cite{guo2018robust_preprint}. Table \ref{tab:datasets} mentions details about the datasets. Standard pre-processing steps—such as text normalization, entity extraction, and format standardization were applied for consistency across datasets. Wikipedia dumps were used as the knowledge base for candidate generation, providing an up-to-date and extensive entity source. 

\begin{table}[h!]
  \centering
  \begin{tabular}{llc}
    \hline
    \textbf{Dataset} & \textbf{Source} & \textbf{Mentions} \\
    \hline
    ACE2004 & News articles & 259 \\
    AIDA    & Reuters news articles & 4464 \\
    AQUAINT & Newswire articles & 743 \\
    MSNBC   & News articles & 656 \\
    CWEB    & Web Pages & 6821 \\
    WIKI    & Wikipedia articles & 11154 \\
    \hline
  \end{tabular}
  \caption{Summary of entity linking test datasets used in our evaluation.}
  \label{tab:datasets}
\end{table}

\renewcommand{\arraystretch}{1.5}  



\subsection{Entity Linking Evaluation Metric}


Disambiguation Accuracy is defined as the following to evaluate our entity linking system

\begin{equation}
  \mathrm{Accuracy} \;=\;
  \frac{\mathrm{TP}_{\text{link}}}
       {\mathrm{TP}_{\text{link}} + \mathrm{FP}_{\text{link}} + \mathrm{FN}_{\text{link}}}
\end{equation}
where $\mathrm{TP}_{\text{link}}$ is the number of mentions correctly linked to their gold entity, $\mathrm{FP}_{\text{link}}$ is the number of mentions linked to an incorrect entity, $\mathrm{FN}_{\text{link}}$ is the number of mentions that should have been linked but were not.

\subsection{Entity Linking Baselines}

As our primary external baseline, we compared with the state-of-the-art ReFinED model \cite{ayoola-etal-2022-refined}. After routing, mentions classified as easy are resolved by ReFinED, which leverages fine-grained entity types and description-based ranking to disambiguate entities without additional reasoning. For secondary analysis, we included the full prompting baseline, which applies LLM-based reasoning to every mention without routing. 

\begin{table*}[htpb]
  \centering
  \resizebox{\textwidth}{!}{
  \begin{tabular}{lcccccc}
    \hline
    \textbf{Method} & \textbf{ACE2004} & \textbf{AQUAINT} & \textbf{AIDA} & \textbf{MSNBC} & \textbf{CWEB} & \textbf{WIKI} \\
    \hline
    ReFinED & 83.98\% & 86.25\% & 83.07\% & 85.98\% & 69.54\% & 85.98\% \\ \hline
    Ours (ReFinEd + Claude 3 Haiku) & \textbf{87.50\%} & \textbf{86.38\%} & 82.10\% & 90.11\% & \textbf{71.63\%} & 80.44\% \\ 
    Ours (ReFinEd + Claude 3.5 Sonnet) & 86.33\% & 83.88\% & \textbf{85.53\%} & 89.96\% & 70.37\% & \textbf{83.61\%} \\
    Ours (ReFinEd + GPT-4.1) & 87.16\% & 84.84\% & 83.64\% & \textbf{90.45}\% & 70.95\% & 83.46\% \\
    Ours (ReFinEd + DeepSeek R1) & 85.58\% & 78.56\% & 83.61\% & 87.82\% & 68.62\% & 82.33\% \\
    \hline
    Full Prompting (Claude 3 Haiku) & \underline{88.28\%} & \underline{87.36\%} & 79.60\% & \underline{90.86\%} & \underline{72.79\%} & 77.72\% \\
    Full Prompting (Claude 3.5 Sonnet) & 86.33\% & 83.19\% & \underline{85.81\%} & 90.25\% & 70.70\% & \underline{83.45\%} \\
    \hline
    Gap (Our Best - ReFinED) & +3.52\% & +0.13\% & +2.46\% & +4.47\% & +2.09\% & -2.37\% \\
    Gap (Our Best - Full Prompting Best) & -0.78\% & -0.98\% & -0.28\% & -0.41\% & -1.16\% & 0.16\% \\

    \hline
  \end{tabular}
  }
  \caption{Accuracy comparison across six entity linking benchmarks. Bolded values indicate the best performance among our approaches, underlined values indicate the best numbers among full prompting. 
  \textbf{\textit{Takeaway:}} \methodAbbrev outperforms ReFinED and achieves results comparable to full prompting pipeline. Breakdown of accuracy on easy and hard cases can be found in Table~\ref{tab:accuracy-breakdown} (Appendix~\ref{sec:appendix}).
}
  \label{tab:hybrid-comparison}
\end{table*}



\section{Results}

\subsection{Router Performance}
We select an optimal decision threshold for the router ($\tau$ = 0.735) on the AIDA validation set (4,784 instances) by maximizing Youden's J statistic. 
The classifier achieves an average AUC of 73.8, accuracy of 65.1\%, F1 score of 75.3, and an average “easy” subset accuracy of 89.3\% across the six test datasets. More details about the Router performance is reported in Table~\ref{tab:router_metrics} in Appendix~\ref{sec:appendix}. Note that out setup prioritizes “easy” subset accuracy over “hard” subset accuracy. The reason is that misrouting a hard example into the easy path is very likely to cause the RefinED model to output an incorrect entity, whereas sending an easy example into the hard path only slightly reduces efficiency while still producing a correct prediction. 


\begin{table*}[htpb]
  \centering
  \resizebox{\textwidth}{!}{
  \begin{tabular}{lcccccc}
    \hline
    \textbf{Mention Type} & \textbf{ACE2004} & \textbf{AQUAINT} & \textbf{AIDA} & \textbf{MSNBC} & \textbf{CWEB} & \textbf{WIKI} \\
    \hline
    Easy Mentions & 129 (50.4\%) & 411 (57.1\%) & 2967 (67.0\%) & 390 (59.5\%) & 6592 (59.3\%) & 4555 (67.3\%) \\
    Hard Mentions & 127 (49.6\%) & 309 (42.9\%) & 1458 (33.0\%) & 266 (40.5\%) & 4530 (40.7\%) & 2213 (32.7\%) \\
    \hline
  \end{tabular}
  }
  \caption{
    Mention distribution across six benchmark datasets after classification by the router. 
    Each cell shows the number of mentions and the corresponding percentage relative to the dataset total. \textbf{\textit{Takeaway:}} More than half of samples across all datasets are resolved using ReFinED, thus saving on expensive LLM reasoning.
  }
  \label{tab:mention-distribution}
\end{table*}

\begin{table*}[htpb]
  \centering
  \resizebox{\textwidth}{!}{
  \begin{tabular}{llrrrrrr}
    \hline
    \textbf{Router Setting} & \textbf{Type} & \textbf{ACE2004} & \textbf{AQUAINT} & \textbf{AIDA} & \textbf{MSNBC} & \textbf{CWEB} & \textbf{WIKI} \\
    \hline
    \textbf{With Router}     & Input  & 268{,}576 & 662{,}139 & 3{,}621{,}636 & 584{,}071 & 10{,}126{,}537 & 4{,}799{,}149 \\
                             & Output & 12{,}811  & 32{,}050  & 106{,}009     & 27{,}552  & 461{,}227     & 157{,}559 \\
    \textbf{Without Router}  & Input  & 519{,}186 & 1{,}428{,}343 & 10{,}387{,}291 & 1{,}322{,}745 & 23{,}275{,}390 & 13{,}384{,}407 \\
                             & Output & 25{,}123  & 72{,}651  & 395{,}849     & 57{,}042  & 1{,}096{,}601 & 488{,}213 \\
    \hline
    \textbf{Input Reduction (\%)}  &        & 48.28\% & 53.63\% & 65.14\% & 55.84\% & 56.49\% & 64.15\% \\
    \textbf{Output Reduction (\%)} &        & 49.01\% & 55.88\% & 73.22\% & 51.70\% & 57.94\% & 67.73\% \\
    \hline
  \end{tabular}
  }
  \caption{
    Input and output token usage for reasoning module across datasets with and without router. 
    \textbf{\textit{Takeaway:}} The router significantly reduces both input and output tokens across all datasets, leading to substantial inference cost savings—especially since output tokens are typically more expensive. 
    \textbf{\textit{Explanation:}} The \textit{without router} rows reflect token usage when all mentions (easy and hard) are processed by the LLM. The \textit{with router} rows include only hard mentions that require LLM reasoning, as easy mentions are handled by ReFinED without incurring LLM tokens.
    }
  \label{tab:router-token-breakdown}
\end{table*}

\subsection{Result Analysis and Discussion}
In this section, we analyze and discuss the results from the angle of questions outlined in the Introduction. 

\textit{1. Can the strategic application of LLM-based techniques within \methodAbbrev significantly improve EL performance for ambiguous mentions (i.e. “hard cases” where traditional models typically falter)?} 
    
Yes. Table~\ref{tab:hybrid-comparison} shows that our hybrid method consistently performs better than the ReFinED baseline on most datasets, and delivers results close to the best full prompting approaches. For example, we improve over ReFinED by +3.52\% on ACE2004, +4.47\% on MSNBC, +2.46\% on AIDA, +2.09\% on CWEB. On AQUAINT, our performance is slightly lower by 0.13\%.

\textbf{Comments on WIKI dataset} We would like to especially point out that it is expected that ReFinED outperforms all other methods on the WIKI dataset. This aligns with the fact that ReFinED was trained specifically on the full English Wikipedia, giving it a natural advantage on this dataset. The WIKI dataset is the only case where ReFinED outperforms \methodAbbrev. 

\textit{2. Do all mentions require LLM reasoning? If not, which mentions can be handled by traditional models? How can we reduce the cost of LLM-based reasoning without sacrificing accuracy?} 

    Our router identifies easy mentions that can be resolved accurately without LLMs. As shown in Table~\ref{tab:mention-distribution}, between 50.4\% to 67.3\% of mentions in each dataset fall into this category. These are processed by ReFinED, demonstrating that a majority of mentions do not require costly LLM inference. This insight enables efficient deployment without degrading performance. 

    \methodAbbrev achieves accuracy that is comparable to full LLM prompting. This highlights that our reasoning module, when applied only to challenging mentions, is not just cost-efficient but also highly effective. 
    More details about model accuracy and inference cost across several LLM-based configurations in Appendix~\ref{sec:appendix}.

\textit{3. How does \methodAbbrev compare to the full prompting approach in terms of accuracy and computational efficiency?} 

    When compared with full prompting (which sends every mention to a reasoning LLM), our method achieves similar accuracy with much less computation. The performance gap is small: within 1\% on most datasets. As shown in Table~\ref{tab:mention-distribution}, hard mentions make up just 32.7 to 49.6\% of the data depending on the dataset. This selective use of LLMs leads to large efficiency gains without compromising accuracy.


\section{Deployment Efficiency}

Modern entity linking systems, such as OneNet \cite{liu-etal-2024-onenet}, use LLMs with reasoning strategies like Chain-of-Thought for every mention. However, not all mentions require this level of reasoning. Our method introduces a router that identifies "easy" mentions and handles them with a lightweight model like ReFinED, skipping LLM reasoning when unnecessary.

To benchmark efficiency, the number of LLM tokens used is tracked and measured using the \texttt{tiktoken} library with the \texttt{cl100k\_base} encoding, which is the same tokenizer employed by ChatGPT-3.5 and GPT-4-turbo. Although originally designed for OpenAI models, this tokenizer serves as a reasonable proxy for Claude models, which do not provide a public tokenizer. Tokens generated by LLaMA 3.1-8B-Instruct (used in the routing module) are excluded from the deployment-efficiency calculation due to their negligible cost (approximately \$0.05 per 1M input tokens and \$0.08 per 1M output tokens). An estimated cost for Router feature generation using LLaMA 3.1-8B-Instruct is provided in Appendix~\ref{appn:router_cost}.

As shown in Table~\ref{tab:mention-distribution}, between 50.4\% and 67.3\% of mentions in public datasets are classified as easy and can be accurately resolved without invoking an LLM. This routing strategy results in an average reduction of 58.25\% in LLM token usage across datasets (Table~\ref{tab:router-token-breakdown}). More importantly, output token usage, which typically incurs higher cost, decreases by an average of 59.25\%, substantially reducing inference cost.



\section{Conclusion}
\methodAbbrev achieves strong entity linking performance by adaptively routing mentions based on contextual difficulty. Compared to ReFinED, it improves accuracy by up to +4.47\%, with an average gain of +2.53\% across datasets. While maintaining comparable performance to full LLM-based prompting methods (within ±1\% on most datasets), \methodAbbrev reduces LLM token usage by an average of 58.25\%, demonstrating significantly better efficiency during deployment.

Beyond benchmarks, \methodAbbrev addresses practical deployment needs: it avoids domain-specific fine-tuning, routes most mentions through a fast path to prevent latency spikes and cost surges, and combines lightweight models with LLM reasoning to ensure robustness under domain shifts and knowledge base updates.

\section*{Limitations}

 
Our study has several limitations. First, due to computational and budget constraints, we were unable to benchmark a broader range of large language models in full prompting settings. Second, we have not yet explored the use of LLMs for direct entity candidate generation; incorporating this could help us better understand the contribution of generation versus reranking through ablation studies. Third, while our routing mechanism shows promising efficiency gains, we have not systematically studied the effect of different router parameter configurations on accuracy. We leave these directions for future work.

\section*{Acknowledgments}
The authors would like to thank Professor Andrew McCallum, Wenlong Zhao, and Wanyong Feng for their mentorship and constructive feedback. We also thank Apple and the University of Massachusetts Amherst Industry Mentorship Program for their collaboration and support throughout this project, as well as the anonymous reviewers for their valuable comments.

\bibliography{custom}

\appendix
\section{Experimental Supplement}\label{sec:appendix}
\subsection{Router Performance Metrics}

We tabulate the full per‐dataset router metrics. Our analysis demonstrates that the router’s accuracy on the subset it labels as “easy” is consistently and substantially higher than on the subset it labels as “hard” (Table \ref{tab:router_metrics}) . This outcome directly reflects our design objective of minimizing false-easy errors—i.e. hard examples erroneously passed to the lightweight ReFined module—because any instance classified as “hard” is instead routed to the reasoning LLM and resolved correctly. In contrast, false-hard errors (truly easy examples sent to the reasoning path) incur only modest extra computation while still yielding accurate entity links.

\begin{table}[htpb]
  \centering
  \scriptsize
  \setlength{\tabcolsep}{6pt}
  \begin{tabular}{lccccc}
    \toprule
    \textbf{Dataset}   
      & \textbf{AUC}   
      & \textbf{Accuracy}   
      & \textbf{F1}    
      & \textbf{Easy-Acc} 
      & \textbf{Hard-Acc} \\
    \midrule
    ACE2004   & \textbf{0.826} & 0.602 & 0.703 & \textbf{0.938}    & 0.260    \\
    AIDA      & 0.782 & \textbf{0.727} & \textbf{0.818} & 0.916    & 0.342    \\
    AQUAINT   & 0.670 & 0.600 & 0.721 & 0.905    & 0.194    \\
    MSNBC     & 0.739 & 0.646 & 0.757 & 0.926    & 0.237    \\
    CWEB      & 0.703 & 0.650 & 0.728 & 0.791    & \textbf{0.444}    \\
    WIKI      & 0.710 & 0.683 & 0.789 & 0.880    & 0.277    \\
    \bottomrule
  \end{tabular}
  \caption{Router performance by dataset. “Easy-Acc” is accuracy on the instances the router predicts as “easy”; “Hard-Acc” is accuracy on the instances the router predicts as “hard”.}
  \label{tab:router_metrics}
\end{table}

\subsubsection{Feature and Ablation Analysis}
\paragraph{Feature importance.}
On the development set, Random Forest impurity reductions rank the features as follows:
the average of $\theta_3$ across candidates is most influential, followed closely by the
average of $\theta_2$. Next comes the entropy of the (base-2) normalized flattened
candidate–score distribution and the margin between the top two penalized candidates
(\texttt{top1} $-$ \texttt{top2}). Sentence length and the average of
$\theta_1$ contribute comparably, while absolute peak scores (\texttt{top1}, \texttt{top2})
provide additional but secondary signal. The average of the LLM-driven score $\phi$
(\texttt{score\_4}) has a smaller yet non-negligible effect, and the sheer number of
candidates (\texttt{n\_cands}) is least informative. As usual, importances are
setup-dependent (e.g., router thresholds, class rebalancing, feature scaling) and are
interpreted as relative contributions rather than absolute effects.

\paragraph{Ablation.}
We ablate features by removing them from the classifier. Eliminating the LLM-driven score $\phi$ (\texttt{score\_4}) changes accuracy by $-1.0\%$ on the \texttt{easy} subset and $+1.0\%$ on the \texttt{hard} subset; because our routing objective prioritizes \texttt{easy} accuracy to avoid unnecessary expensive reasoning, and because $\phi$ is computed by a lightweight, fast, low-cost scorer, we retain $\phi$ in the full model. Removing high-importance signals (avg~$\theta_{3}$, avg~$\theta_{2}$) degrades overall accuracy more substantially, confirming their central role. The exact trade-offs depend on router configuration; a more comprehensive exploration of router design is left to future work.

\subsection{Cost for Router LLM-based Features}
\label{appn:router_cost}

We estimate the cost for generating the LLM score used as one of the features in Router training. We compute the total input and output tokens consumed for generating the feature for all datasets under consideration. We use the API pricing as per Groq\footnote{\url{https://groq.com/pricing}} as an estimate for running the LLM inference (\$0.05/M for input tokens and \$0.08/M for output tokens). Table \ref{tab:cost-router} shows the number of input and output tokens (in  millions) and the total cost incurred. Despite the high token usage, running inference using a 8B model is cheaper than using other proprietary models, thus making the router training resource efficient.

\begin{table}[htbp]
  \centering
  \scriptsize
  \setlength\tabcolsep{3pt}
  \renewcommand{\arraystretch}{1.1}
  \resizebox{1.0\columnwidth}{!}{%
    \begin{tabular}{l c c c}
      \toprule
      \textbf{Dataset} & \textbf{Input Tokens (M)} & \textbf{Output Tokens (M)} & \textbf{Cost (\$)} \\
      \midrule
      ACE2004  & 0.421 & 0.027 & 0.023 \\
      AQUAINT  & 1.142 & 0.084 & 0.064 \\
      MSNBC    & 1.174 & 0.071 & 0.064 \\
      AIDA     & 17.581 & 0.539 & 0.922 \\ 
      CWEB     & 24.154 & 1.398 & 1.320 \\
      WIKI     & 10.814 & 0.677 & 0.595 \\
      \midrule
      \text{Total} & \text{55.29} & \text{2.80} & \text{2.99} \\
      \bottomrule
    \end{tabular}%
  }
  \caption{Estimated cost for generating LLM‐based features using observed token usage}
  \label{tab:cost-router}
\end{table}


\subsection{Prompting Strategy Comparison}

We evaluated several prompting strategies for entity linking on the ACE2004 hard cases dataset using the Haiku 3 model. 

Table~\ref{tab:prompt-strategy-accuracy} compares different prompting strategies on ACE2004 hard cases. Zero-shot prompting relies on the model's prior knowledge without examples, achieving 75.89\% accuracy. Few-shot provides illustrative examples, while Chain-of-Thought (CoT) encourages step-by-step reasoning. Combining few-shot with CoT yields the best performance (81.3\%), as it guides the model with both examples and structured reasoning. Contrastive prompting exposes correct and incorrect examples to reduce common errors (79.46\%), and ReAct-style prompting interleaves reasoning with actions, which is less beneficial for this self-contained task (76.79\%).

Table~\ref{tab:prompt-strategy-accuracy} summarizes the accuracy achieved by each strategy.

\begin{table}[htbp]
    \centering
    \small
    \begin{tabular}{lc}
        \toprule
        \textbf{Strategy} & \textbf{Accuracy (\%)} \\
        \midrule
        Zero-shot             & 75.89 \\
        Few-shot         & 76.79 \\
        Few-shot + CoT        & \textbf{81.30} \\
        Contrastive           & 79.46 \\
        REACT                 & 76.79 \\
        \bottomrule
    \end{tabular}
    \caption{Accuracy of different prompting strategies on the ACE2004 hard cases dataset.}
    \label{tab:prompt-strategy-accuracy}
\end{table}

\subsection{Cost-Performance Analysis}

The trade-off between model accuracy and inference cost is analyzed across several LLM-based configurations. Based on a total usage of 20.06 million input tokens and 0.80 million output tokens, cost estimates are computed using publicly available API pricing for each model.

\begin{figure}[ht]
    \centering
    \includegraphics[width=0.48\textwidth]{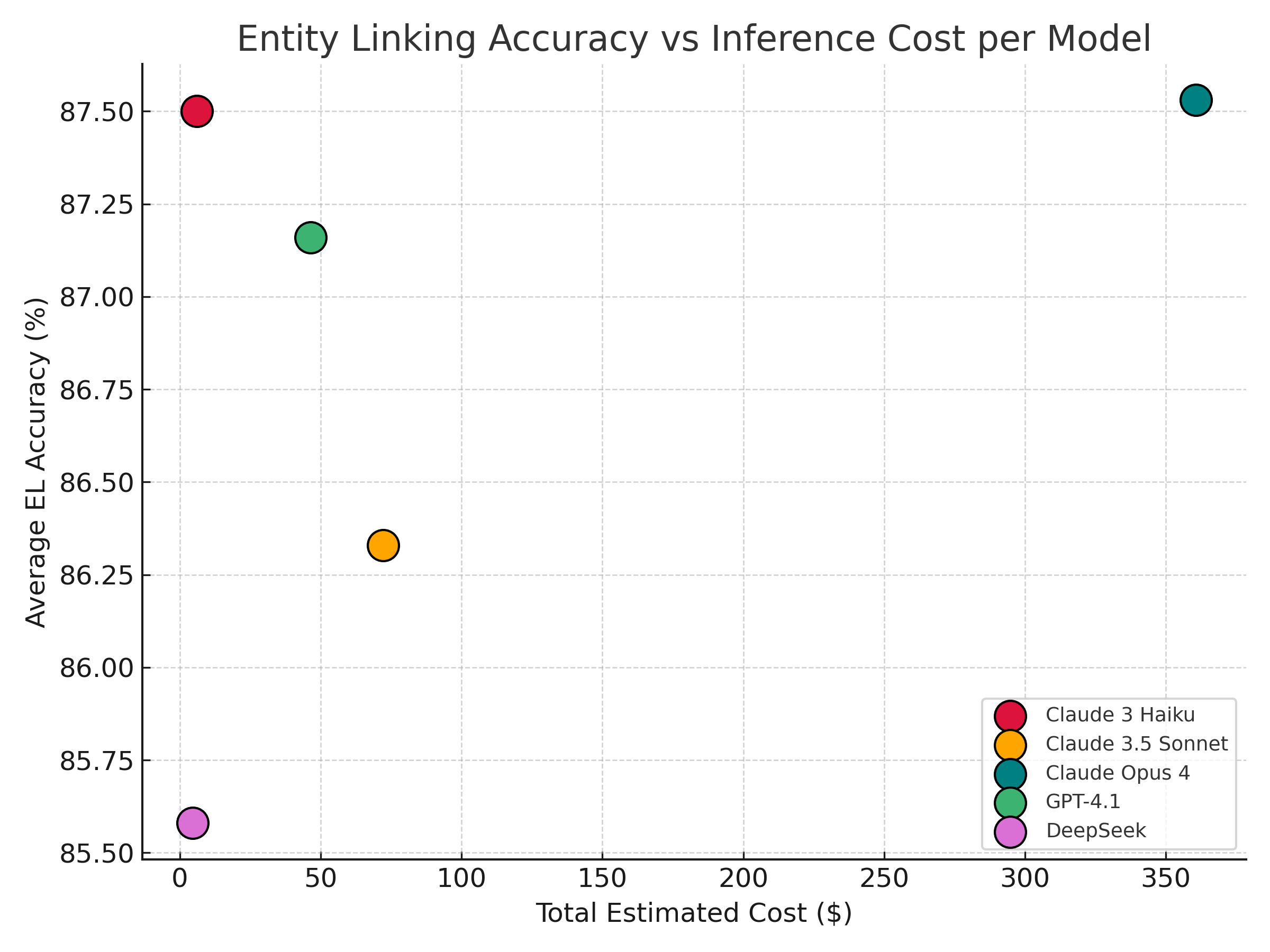}
    \caption{Entity linking accuracy vs. total inference cost across LLM-based configurations.}
    \label{fig:cost-performance}
\end{figure}

\begin{table}[htbp]
    \centering
    \small
    \begin{tabular}{lccc}
        \toprule
        \textbf{Model} & \textbf{Input} & \textbf{Output} & \textbf{Cost (\$)} \\
        \midrule
        Claude 3 Haiku      & \$0.25/M & \$1.25/M  & 6.01 \\
        Claude 3.5 Sonnet   & \$3.00/M & \$15.00/M & 72.14 \\
        Claude Opus 4       & \$15.00/M & \$75.00/M & 360.72 \\
        GPT-4.1             & \$2.00/M & \$8.00/M  & 46.50 \\
        DeepSeek            & \$0.14/M & \$2.19/M  & 4.55 \\
        \bottomrule
    \end{tabular}
    \caption{Estimated inference cost using observed token usage (20.06M input, 0.80M output) and current API pricing. 
    }
    \label{tab:cost-performance}
\end{table}

As shown in Table~\ref{tab:cost-performance} and Figure~\ref{fig:cost-performance}, Claude 3 Haiku provides the best cost-effectiveness, achieving 87.5\% accuracy at only \$6.01. DeepSeek offers the lowest overall cost (\$4.55) with slightly lower performance (85.58\%). In contrast, premium models such as Claude Opus 4 and GPT-4.1 offer marginal accuracy improvements at significantly higher cost. These results highlight the importance of model choice in reducing cost and achieving efficiency gains. 

\subsection{Accuracy Breakdown on Easy and Hard Cases}

Entity linking accuracy is reported separately for easy and hard mentions to provide a clearer evaluation of the effectiveness of selective LLM reasoning.

Table~\ref{tab:accuracy-breakdown} summarizes accuracy across six benchmark datasets. Results are presented for ReFinED alone, the proposed hybrid reasoning variants, and full prompting using Claude models. The rows are organized into three categories: (1) models applied to easy cases, (2) models applied to hard cases, and (3) full prompting baselines for comparison.

\begin{table*}[htbp]
\centering
\small
\begin{tabular}{lcccccc}
\toprule
\textbf{Model} & \textbf{ACE2004} & \textbf{AQUAINT} & \textbf{AIDA} & \textbf{MSNBC} & \textbf{CWEB} & \textbf{WIKI} \\
\midrule
\multicolumn{7}{l}{\textit{Easy Cases}} \\
\midrule
ReFinEd Only & 93.80\% & 90.50\% & 91.60\% & 92.60\% & 79.10\% & 88.00\% \\
Reasoning (Claude 3 Haiku) & 95.35\% & 92.21\% & 87.87\% & 93.85\% & 81.07\% & 83.95\% \\
Reasoning (Claude 3.5 Sonnet) & 93.80\% & 89.29\% & 92.01\% & 93.08\% & 79.66\% & 87.73\% \\
\midrule
\multicolumn{7}{l}{\textit{Hard Cases}} \\
\midrule
Reasoning (Claude 3 Haiku) & 81.10\% & 80.91\% & 62.76\% & 86.47\% & 60.75\% & 64.89\% \\
Reasoning (Claude 3.5 Sonnet) & 78.74\% & 75.08\% & 73.18\% & 86.09\% & 57.66\% & 74.56\% \\
\midrule
\multicolumn{7}{l}{\textit{Full Prompting Baselines (All Cases)}} \\
\midrule
Claude 3 Haiku & 88.28\% & 87.63\% & 79.60\% & 90.86\% & 72.79\% & 77.72\% \\
Claude 3.5 Sonnet & 86.33\% & 83.19\% & 85.81\% & 90.25\% & 70.70\% & 83.42\% \\
\midrule
\multicolumn{7}{l}{\textit{Our Hybrid Accuracy (ReFinEd for Easy Cases + LLM for Hard Cases)}} \\
\midrule
Claude 3 Haiku & 87.50\% & 86.38\% & 82.10\% & 90.11\% & 71.63\% & 80.44\% \\
Claude 3.5 Sonnet & 86.33\% & 83.88\% & 85.53\% & 89.96\% & 70.37\% & 83.61\% \\
\midrule
ReFinEd Baseline (All Cases)  & 83.98\% & 86.25\% & 83.07\% & 85.98\% & 69.54\% & 85.98\% \\
\bottomrule
\end{tabular}
\caption{Entity linking accuracy (\%) across datasets, separated by case difficulty and reasoning strategy.}
\label{tab:accuracy-breakdown}
\end{table*}

\subsection{Prompting details}

Table~\ref{tab:module_prompts_combined} lists the prompts used for the Router score computation and the reasoning module.

\begin{table*}[htpb]
  \centering
  \renewcommand{\arraystretch}{1.0}
  \setlength{\tabcolsep}{5pt}
  \footnotesize
  \resizebox{\textwidth}{!}{%
  \begin{tabular}{@{} l p{0.72\textwidth} @{}}
    \toprule
    \textbf{Module} & \textbf{Prompt} \\
    \midrule
    LLM scores (confidence) 
        & \textbf{System Instruction:} \newline
          Context: <context> \newline
          Mention: <mention> \newline
          Candidate: <candidate> \newline
          On a scale from 0 to 1, how confident are you that this candidate correctly resolves the mention? Reply with a single number between 0.0 and 1.0. \newline
          \textbf{Output Format:} \newline
          \texttt{\{"scores": \{"<candidate\_id>": <confidence>\}\}} \\
    \addlinespace[4pt]

    Entity Linking (All datasets)
        & \textbf{System Instruction:} \newline
          You are an expert in entity linking. Given a mention, its context, and a list of candidates (each with title, description, and ID), identify the most relevant entity. Pay close attention to the entity descriptions as they provide crucial information for disambiguation. If no candidate is suitable, return -1. Only return the answer in the specified JSON format. \newline
          \textbf{Examples with reasoning:} \newline
          \textbf{Example 1:} Mention: "Apple" \newline
          Context: "I love my new Apple iPhone." \newline
          Candidates: 1. Apple Inc. — Technology company known for iPhone and iPad [Q312], 2. Apple (fruit) — Edible fruit [Q89], 3. Apple Records — British record label [Q213710] \newline
          Reasoning: The context mentions 'iPhone'; the correct entity is Apple Inc. \newline
          Output: \texttt{\{"linked\_entity": 1, "entity\_id": "Q312", "entity\_title": "Apple Inc.", "reasoning": "..."\}} \newline
          \textbf{Example 2:} Mention: "Paris" \newline
          Context: "I'm planning a trip to Paris next summer." \newline
          Candidates: 1. Paris (France) [Q90], 2. Paris (mythology) [Q167646], 3. Paris, Texas [Q43668] \newline
          Reasoning: 'trip to' implies travel to a city; correct entity is Paris, France. \newline
          Output: \texttt{\{"linked\_entity": 1, "entity\_id": "Q90", "entity\_title": "Paris", "reasoning": "..."\}} \newline
          \textbf{Example 3:} Mention: "XYZ" \newline
          Context: "The company XYZ is not well known." \newline
          Candidates: 1. Microsoft [Q2283], 2. Google [Q95], 3. Amazon [Q3884] \newline
          Reasoning: 'XYZ' does not match any candidate. \newline
          Output: \texttt{\{"linked\_entity": -1, "entity\_id": "-1", "entity\_title": "None", "reasoning": "..."\}} \newline
          \textbf{Example 4:} Mention: "Einstein" \newline
          Context: "Einstein's theory of relativity revolutionized physics." \newline
          Candidates: 1. Albert Einstein [Q937], 2. Einstein (band) [Q12309581], 3. Einstein (disambiguation) [Q214395] \newline
          Reasoning: Context refers to physics; correct entity is Albert Einstein. \newline
          Output: \texttt{\{"linked\_entity": 1, "entity\_id": "Q937", "entity\_title": "Albert Einstein", "reasoning": "..."\}} \newline
          \textbf{Example 5:} Mention: "Java" \newline
          Context: "I'm learning Java programming language for software development." \newline
          Candidates: 1. Java (programming language) [Q251], 2. Java (island) [Q252], 3. Java (coffee) [Q2642722] \newline
          Reasoning: Context refers to programming; correct entity is Java (programming language). \newline
          Output: \texttt{\{"linked\_entity": 1, "entity\_id": "Q251", "entity\_title": "Java (programming language)", "reasoning": "..."\}} \\
    \bottomrule
  \end{tabular}%
  }
  \caption{Prompts used by each module in our pipeline.}
  \label{tab:module_prompts_combined}
\end{table*}

\end{document}